\documentclass[12pt]{report}
\usepackage[utf8]{inputenc}
\usepackage[a4paper,width = 150mm, top = 25mm, bottom = 25mm]{geometry}
\usepackage{graphicx}
\graphicspath{{Images_and_Figures/}}
\usepackage[export]{adjustbox}
\usepackage{float}
\usepackage{indentfirst}
\usepackage[sorting = none]{biblatex}
\usepackage{dirtytalk}
\usepackage{algorithm} 
\usepackage{algpseudocode} 
\usepackage{amsmath}
\usepackage{multirow}
\usepackage{multicol}
\usepackage{hyperref}
\hypersetup{
    colorlinks=true,
    linkcolor=blue,
    filecolor=magenta,      
    urlcolor=blue,
    citecolor=blue,
}
\title{
    {A novel k-means clustering approach using two distance measures for Gaussian data}  \\
    {\small School of Mathematical Sciences}\\
    {\small College of Science} \\
    {\small Rochester Institute of Technology} \\
    {\small Advisor/Supervisor: Dr. Nathan Cahill}
    }

\author{Naitik H. Gada}
\date{$22^{nd}$ November, 2021}

\begin{document}

\maketitle

\chapter*{Abstract}
Clustering algorithms have long been the topic of research, representing the more popular side of unsupervised learning. Since clustering analysis is one of the best ways to find some clarity and structure within raw data, this paper explores a novel approach to \textit{k}-means clustering. Here we present a \textit{k}-means clustering algorithm that takes both the within cluster distance (WCD) and the inter cluster distance (ICD) as the distance metric to cluster the data into \emph{k} clusters pre-determined by the Calinski-Harabasz criterion in order to provide a more robust output for the clustering analysis. The idea with this approach is that by including both the measurement metrics, the convergence of the data into their clusters becomes solidified and more robust. We run the algorithm with some synthetically produced data and also some benchmark data sets obtained from the UCI repository. The results show that the convergence of the data into their respective clusters is more accurate by using both WCD and ICD measurement metrics. The algorithm is also better at clustering the outliers into their true clusters as opposed to the traditional \textit{k} means method. We also address some interesting possible research topics that reveal themselves as we answer the questions we initially set out to address. 

\tableofcontents

\chapter{Introduction}
\par Machine learning can be broadly classified into three categories of algorithmic techniques depending on the type of problem being faced. These techniques also vary heavily, based on the learning algorithms themselves. They are: Supervised Learning, Reinforcement Learning and Unsupervised Learning. 

\par Supervised learning makes use of predetermined classes of the responses to train the underlying algorithm and make predictions based on the same. The features or the independent variables, \textbf{\textit{X}} are mapped to the responses or the dependent variables, \textbf{\textit{Y}}. The algorithm learns from this mapping to make predictions on unseen data. Hence the name, machine learning \cite{24}. It is the human equivalent of learning from our past experiences in order to gather knowledge that is imperative to improve our ability to perform future tasks in the real world. Supervised learning algorithms can either be categorized as regression or classification. Dependent variables with continuous data fall into the regression category whereas dependent variables with discrete data labels fall into the classification category \cite{22}. 

\par Reinforcement learning is another entity altogether wherein the algorithm interacts with the environment based on some actions, \textbf{\textit{A}} = ${a_1, a_2, ... a_N}$ and tries to produce an optimal policy or decision, by trial and error, based on rewards (or punishments), \textbf{\textit{R}} = ${r_1, r_2, ... r_N}$. The goal with this algorithm is to produce a decision based on a path that maximizes the rewards or minimize the punishments by the time the stopping criteria is met \cite{27}, \cite{29}.

\par Contrary to the supervised learning algorithms where the output data is labelled, unsupervised algorithms almost seem black boxed because there is no coherent knowledge about the data prior to processing it \cite{18}. The features or the independent variables, \textbf{\textit{X}} = ${x_1, x_2, .... x_N}$, where \textbf{\textit{X}} is an N-dimensional set of vectors, have no associated labels and are hence at the mercy of the processing algorithm to identify the homogeneity within the data set. This data could be customer information at a marketing research company, a set of handwritten alphabets or numbers, confidential patient data for the medical records, pixel information from an image. Unsupervised learning and specifically exploratory data analysis is now being widely used as the first step in gaining a deeper understanding of the raw data \cite{26}. One approach calls for the use of metaheuristic algorithms to deal with the NP-hardness of the unsupervised learning algorithms \cite{19}. Metaheuristic algorithms are high level guiding strategies that search the solution space for the most optimal solution. Despite being able to reduce the time complexity of finding a solution, metaheuristic algorithms have one major drawback: they are known to get stuck in the local optima \cite{19}, \cite{28}. Another approach calls for genetic algorithm to better search the solution space for optimized results \cite{11}. Genetic algorithms are a part of the \say{Evolutionary Algorithms} that mimic real life evolution to search for an optimized solution. Other evolutionary algorithms include Pattern Search, Particle Swarm Optimization, Simulated Annealing \cite{11}, \cite{12}, \cite{19}. Whatever the approach may be, in order to solve the unsupervised learning problems, we have to deploy methods that find patterns beyond the raw data which is nothing but unstructured noise. One method to understand the homogeneity in the underlying raw data is clustering analysis \cite{1}, \cite{23}, \cite{25}. 

\section{Clustering Analysis}
Clustering analysis is a class of unsupervised learning techniques that is the most widely researched \cite{1}, \cite{10}, \cite{21}. It garners more interest particularly because of its applications in deep learning and its potential ability to produce sophisticated artificial intelligence algorithms \cite{1}. Unsupervised learning algorithms, specifically clustering algorithms have been researched for the last four decades \cite{22}. The goal of clustering is to separate the unlabelled data into a finite set of \say{clusters} with similar properties among the data within the same clusters but differing properties from the data in other clusters \cite{2}, \cite{4}. Its applications include, but are not limited to data mining, image segmentation, vector quantization, pattern recognition, etc. \cite{2}. The definition of a \say{good} cluster remains implicit in that it depends on the application. There are many methods to classify these clusters, both ad hoc and systematic \cite{2}, \cite{4}, \cite{3}. Hierarchical clustering uses recursion to find clusters within the data by either a \say{top down} or a \say{bottom up} approach with a higher than quadratic complexity \cite{7}. This makes them highly unsuitable for larger data sets. In contrast, partitional algorithms find all the clusters within the data simultaneously and are computationally much cheaper than hierarchical algorithms making them favorable for large data sets. Clustering analysis is fundamental in deriving an insight into purely unstructured and noisy data. What sets the tone for further analysis of the clustered data is the similarity in properties of data belonging to the same clusters. These properties can either be distance based measures like Euclidean or Manhattan or they can be density based clustering like DBSCAN and BIRCH \cite{2}. Marketing research companies make use of clustering analysis to better target their audiences \cite{29}. Similarly even social media companies make use of clustering analysis and graph theory as a way to link similar social groups together and target ads based on the same. 
One such popular method of clustering is using a pre-determined number of clusters to classify the data into \textit{k} different clusters \cite{10}. \textit{k}-means clustering is undoubtedly one of the most popular clustering algorithms in existence today evident by the hundreds of publications on the topic in the last 30-40 years, each one extending and improving the \textit{k}-means algorithm in some way or another.

\newpage

\section{\textit{k}-means Clustering}
Given a set of \textit{p} data points in a \textit{d}-dimensional space, $R^d$, the goal of \textit{k}-means clustering is to populate the \textit{p} data points into \textit{k} clusters, where the value of \textit{k} is predetermined \cite{2}. The \say{grouping} of the data points is achieved in a way that minimizes the distance between the cluster centers and their corresponding data points. One of the biggest drawbacks of the \textit{k}-means clustering algorithm is that the number of clusters, \textit{k} have to be predetermined before the algorithm is applied. This may seem like a trivial step since the human eye is able to estimate the number of clusters when the data is first visualized. On the other hand, this task becomes difficult if the data is highly dimensional which makes it hard to visualize. It becomes even trickier to predetermine the number of clusters if the algorithm is to be deployed to multiple data sets at a time. The other major drawback of this algorithm is that it is sensitive to the initial position of the cluster centers which causes it to converge at local minima rather than the global optima \cite{5}, \cite{2}. Despite the uncertainty of achieving a global minima, \textit{k}-means clustering at least ensures convergence. As mentioned earlier, \textit{k}-means is a partitional algorithm which makes it computationally favorable to hierarchical clustering. Albeit as we will find out later in this paper it is not immune to data that is not normal or even extremely large data sets. In this paper, a simple step is proposed to increase the efficiency of the clustering and also to improve the overall accuracy of the algorithm. Instead of using just one distance measure to minimize the homogeneity within a cluster, we propose adding a distance measure to maximize the heterogeneity among different clusters. In other words, by increasing the inter-cluster distance (ICD) while decreasing the within-cluster distance (WCD), it can be hypothesized that the algorithm performs better.

\chapter{Related Work}
\section{Literature Assessment}
\par \textit{k}-means clustering has seen hundreds of iterations over the years, judging by the number of published works you can find \cite{1}, \cite{2}, \cite{4}, \cite{7}, \cite{10}. Every iteration of the research work seeks to improve the algorithm in some way or another. It is well known that \textit{k}-means is based on achieving similar properties within the same cluster. More often that not, the property in question is a distance measure between the cluster center and the corresponding data points. This distance measure is called the mean squared distance and is usually taken to be the Euclidean distance \cite{10}. One paper \cite{30} evaluates the performance of the algorithm with other distance measures like\\

Manhattan distance:
$$ d_{xy} = \sum_{k = 1}^{n} |x_k - y_k| $$

Chebyshev distance:
$$ d_{xy} = max_k(|x_k - y_k|) $$

Minkowski distance:
$$ d_{xy} = \bigg(\sum_{k = 1}^{n} |x_k - y_k|^p \bigg)^{1/p} $$

and concludes that the best performance of the algorithm is achieved by using the Euclidean distance and the worst performance is seen with the Manhattan distance. What it fails to do is provide an accuracy score associated with each distance measure. Sure, it concludes that the Euclidean distance assists the algorithm in providing the best performance but the study fails to clearly articulate the assessment method of the algorithm. Similarly, another study successfully exhibits the performance of the algorithm using an inter-cluster distance management based on centroid estimation but it only relies on synthetic data sets for its assessment \cite{16}. This study includes three different methods for centroid estimation, viz. random selection, mean distance model and inter-cluster distance model. Overall it addresses its hypothesis fairly well and definitely takes an interesting step towards improving the efficiency of the algorithm. Initial starting conditions are mentioned as one of the most prominent issues facing \textit{k}-means clustering. A research study explores eight different initialization methods against thirty two publicly available data sets and concludes that a few of the linear time initialization methods elevate the overall performance of the \textit{k}-means algorithm while keeping its computational complexity in check \cite{7}. Several research methods also address the initialization  problem using metaheuristics and stochastic global optimization methods like simulated annealing and genetic algorithms \cite{4}, \cite{11}, \cite{12}, \cite{19}. There is a similar pattern to most of these approaches and each one tries to improve the algorithm by nuanced changes in the parameters and hyper parameters. Despite these improvements, the highest accuracy score achieved is no more than 80-85 percent. In addition to this low score, none of the methods include a distance measurement to maximize the inter-cluster distance (ICD). A couple of these papers again rely on purely synthetic data for their evaluations. Since the generation of this synthetic data relies on drawing from different distributions, it makes it harder for a fair comparison of the performance of the algorithm \cite{18}. A global \textit{k}-means algorithm approaches the problem of clustering in an incremental method and adds a cluster center at each iteration dynamically through a global search procedure \cite{4}. While the algorithm in this study is quite sound and definitely exhibits its advantages over the normal \textit{k}-means algorithm, it only reports its findings in a metric called \say{clustering error} and does not address the accuracy, sensitivity, specificity or the F1 score. This makes it harder to evaluate the overall performance of the algorithm.
\par Most of the literature research suggested that even though the \textit{k}-means algorithm was improved in some form or another, the corresponding study lacked one or more aspects of the approach undertaken in this paper. The goal of this paper is to evaluate the performance of the proposed algorithm on several benchmark data sets and use the overall accuracy, sensitivity, specificity and F1 score to better quantify that evaluation.

\section{Inspiration and Hypothesis}
\par The main idea that drove this research was the \textit{k}-means clustering approach not found in the literature review. Most of the review papers utilized one or the other measurement metric - either the within cluster distance or the intra-cluster distance. Several studies also explored the initialization problem of \textit{k}-means, some others looked at the selection of \textit{k} and the methods surrounding the most optimal value of \textit{k}. A number of recent research has revolved around the use of metaheuristics in clustering analysis. Some global optimization techniques like simulated annealing or evolutionary algorithms like genetic algorithm, particle swarm optimization have also been the center of \textit{k}-means clustering. With the novel algorithm presented in this project, the assumption is that by using both the within-cluster distance and inter-cluster distance, we could get a more robust idea on the performance of the algorithm as a whole. The hypothesis is that by including the ICD in the algorithm, we'll see higher accuracy, sensitivity, specificity and F1 score.

\chapter{Algorithm Overview}
\section{Description}
\par The basic algorithm clusters raw data into a predetermined number of clusters. The first step of this algorithm is to determine the number of clusters needed. For easily decipherable data, a visualization step is all that is needed to estimate the number of clusters in the data. For larger, more complex data sets, this can be achieved by a number of methods like the Davies Bouldin index, Calinski Harabasz index, Gap Evaluation, Silhouette Evaluation \cite{10}. The most popular is the elbow method that makes use of the distortion or inertia of the entire data set. Distortion is defined to be the average of the squared distances from the cluster centers of the respective clusters and inertia is just the sum of squared distances of the samples to their closest cluster center \cite{3}, \cite{4}, \cite{10}. For this paper we used the Calinski Harabasz index that has an inbuilt function in Matlab and confirmed the validity using the elbow method. 
\par Once the cluster validity was established, the algorithm requires the initialization of the cluster centers. For the initialization step, a random permutation of indices equal to the number of clusters was generated and used to map the initial cluster centers from the given data set. The upper and lower bound of the cluster centers were the maximum and minimum of the data set. The next step was to evaluate the Euclidean distance from the cluster centers generated in the previous step to the rest of the data set and associate a cluster number with each pairing. The association was achieved by taking the lowest distance of a data point to one of the clusters. This was done in a loop for the entire data set and the average of the sum of the within-cluster distance was then taken. The average was taken to be the new cluster center for the next iteration. This was repeated until the cluster centers did not change, which indicated the optimal positions of the clusters. The final step of this algorithm was to compare the association of the data points and their cluster membership to the original data set. Once the comparison was made, the assigned clusters were used to calculate the overall accuracy of the algorithm along with its sensitivity, recall and F1 score. Having these scores along with the confusion matrix gave us a quantifiable insight into its performance.

\section{Terms and Definitions}
A set of \textit{p} data points in a multidimensional space, $R^d$ contain instances and attributes (or features). For the ease of understanding, the data set was thought of as an  \textit{m} x \textit{n} matrix. The total number of elements in this matrix would then be, \textit{m} x \textit{n} = \textit{p}. Instances are the number of observations, \textit{m} (or rows of a matrix) and features are the number of total dimensions, \textit{n} (or columns of a matrix). \textit{k} is the number of clusters in the data set which is predetermined. The initial centroid matrix will be \textit{k} x \textit{n}. The goal of this algorithm is to minimize the within-cluster distance (WCD) and maximize the inter-cluster distance (ICD). WCD is defined as the total distance of a cluster center to each corresponding data point in that cluster, summed over the number of clusters. Mathematically, for \textit{k} clusters:
$$ C_j = \sum_{\overrightarrow{x_i} \in C_j} ||\overrightarrow{x_i} - \overrightarrow{C_j} ||^2 $$

for \textit{j} = \textit{1},..., \textit{k}. \\

ICD is the total distance of a data point belonging to cluster \textit{j} and a data point not belonging to cluster \textit{j}, summed over all clusters:

$$ D_{ip} = \sum_{\overrightarrow{x_i} \in C_i} \sum_{\overrightarrow{x_p} \notin C_j} || \overrightarrow{x_i} - \overrightarrow{x_p}||^2 $$ \\
$$ D = \sum_{i} \sum_{p} D_{ip} $$

The total metric of the algorithm is the relative cost, which is the ratio of WCD and ICD:

$$ \sum_{j = 1}^k \frac{\sum C_j}{D} $$

The overall accuracy of the algorithm is evaluated using a confusion matrix. A confusion matrix is a $\textit{k}^{th}$ order square matrix. It is a matching matrix that compares the actual cluster labels from the data set to the output labels. The elements of the matrix are given by:
$$ \begin{bmatrix}

TP & FN \\
FP & TN

\end{bmatrix} $$

where, TP is True Positive, FN is False Negative, FP is False Positive and TN is True Negative. The overall accuracy (OA) of the algorithm is calculated by adding the diagonal elements and dividing by the total of all the elements of the matrix. That is,
$$ OA = \frac{TP + TN}{TP + FN + FP + TN} $$

Precision, also known as the positive predictive value (PPV) is the ratio of TP and the sum of TP and FP:
$$ PPV = \frac{TP}{TP + FP} $$ 

Recall is the true positive rate (TPR). It's the ratio of the true positives and the total positives:
$$ TPR = \frac{TP}{TP + FN} $$

F1 score is the mean of precision and sensitivity:
$$ F_1 = \frac{2*TP}{2*TP + FP + FN} $$
\\
\\

\section{Pseudocode}

\begin{algorithm}
	\caption{k-means clustering} 
	\begin{algorithmic}[1]
	\State Calculate number of clusters
	\State Initialize cluster centers
		\For {$iteration=1,2,\ldots, Z$}
			\For {$iteration=1,2,\ldots,M$}
				\State Calculate $C_j$,
				\State Compute $D$
				\State Optimize relative cost, $\frac{C_j}{D}$
				\For{$iteration = 1 \ldots,k$}
				    \State mean$(Cluster_j)$
				\EndFor
				\State New centroids = mean$(Cluster_j)$
			\EndFor
			\State Compute the confusion matrix
			\State Calculate OA, Precision, Recall, F1 score
		\EndFor
	\end{algorithmic} 
\end{algorithm}

\chapter{Data}
\section{Synthetic Data}
In order to quantify the proposed algorithm, a couple of synthetic data sets were generated from a normal distribution. One data set had a low variance to make the clustering more apparent and the other had a high variance. The idea with two variances was to generate one data set with more robust clusters and the other data set with more overlapping clusters. Having data sets with different variances would give a more accurate understanding on the performance of the algorithm. More importantly, it would allow for a more clear understanding of the performance of the algorithm against clustering that is not well defined. The intention was to visualize how the clustering changed and how the algorithm picked up outliers in the data set.
\par The first data set was generated with a variance of 0.5 to allow for a higher density within the clusters and the second one was generated with a variance of 1 for relatively ill-defined clusters. In each data set, there are a total of 3 clusters $(k = 3)$. In order to space out the clusters and define the coordinates around which the clusters are centered, a Kronecker tensor product was used to offset each cluster. A Kronecker product is defined as the product of two matrices of arbitrary sizes, resulting in a block matrix. It is denoted by $\otimes$.
Suppose $A$ is a $m \times n$ matrix and $B$ is a $p \times q$ matrix, then the Kronecker product, $A \otimes B$ is a resulting $mp \times nq$ block matrix \cite{31}. Mathematically,
$$ A \otimes B = 
\begin{bmatrix}
a_{11}B & \cdots & a_{1n}B \\
\vdots & \ddots & \vdots \\
a_{m1}B & \cdots & a_{mn}B
\end{bmatrix} $$

The first iteration of generating this synthetic data is run using a 2-dimensional data set and the next iteration is run using a 3-dimensional data set. This gives us a total of 4 synthetic data sets - Two 2-dimensional sets each with a variance of 0.5 and 1, two 3-dimensional sets each with a variance of 0.5 and 1.

\newpage

\section{UCI Machine Learning Repository}
The UCI Machine Learning Repository hosts a collection of databases and data sets that are used by researchers all over the world for the analysis of machine learning algorithms. The advantage of using some of these data sets hosted by the repository is the availability of labelled classes within the data set, which makes it easier to evaluate and quantify the algorithms using the pre-determined labels. As has been mentioned, the raw data in unsupervised learning has no labels and thus makes it hard to evaluate an algorithm against it. Clustering this data provides us no reference into the properties of the different clusters.
\par The first of this data set is the Iris data set. This data set is the most popular amongst researchers in the world of machine learning and pattern recognition. The set contains 150 data points (or instances) and four features (or attributes). The first attribute is the sepal length (in cm), the second is the sepal width (in cm), the third is the petal length (in cm) and lastly the petal width (in cm). The 150 instances are divided into 3 classes of 50 instances each. Each of the three classes represents a type of Iris flower. It is a perfect data set for clustering analysis and the benchmark set to analyze algorithms against \cite{32}. 
\par The second data set is the Wine data set which is a result of chemical analysis of wines that were grown in the same region of Italy but from three different cultivators. It has 178 instances and 13 attributes (Alcohol content, Malic acid content, Ash, Alcalinity of ash, Magnesium, Total phenols, Flavanoids, Nonflavanoid phenols, Proanthocyanins, Color intensity, Hue, OD280/OD315 of diluted wines, Proline) that make up the data. The classification column contains the label of the three cultivators identified by 1, 2, or 3. There are 59 instances of Class 1, 78 instances of Class 2 and 48 instances of Class 3 \cite{32}. 
\par The third and final data set is the Wisconsin Breast Cancer data set that contains features computed from a digitized image of breast mass. The classification is based on whether an image shows a benign mass or a malignant mass (2 classes). It has a total of 569 instances and 9 attributes (not including the ID of the sample and the class) that help classify a data point as being malignant or benign. The features are: clump thickness, uniformity of cell size, uniformity of cell shape, marginal adhesion, single epithelial cell size, bare nuclei, bland chromatin, normal nucleoli, mitoses \cite{32}.

\chapter{Results and Observations}
\section{Synthetic Data Sets}
\subsection{2D Data with variance = 0.5}
The first set of synthetic data that was generated was a 2-dimensional data set with a variance of 0.5.
\begin{figure}[H]
\centering
\includegraphics[scale=0.75]{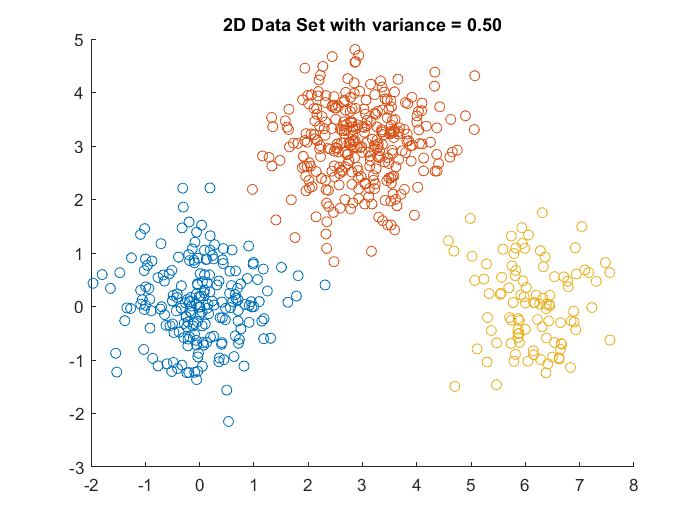}
\caption{2-dimensional data with variance = 0.5}
\label{fig_2D_var_0.5}
\end{figure}

As seen in \textbf{figure \ref{fig_2D_var_0.5}}, the clusters are segregated enough which makes them easy to identify since they are closely packed. The accuracy results of this data set are summarized in \textbf{table \ref{tab_2D_var_0.5}}. 

\begin{table}[H]
\centering
\begin{tabular}[c]{ |p{2.5cm}||p{2.5cm}|p{2cm}|p{2cm}| p{2cm}| }
 \hline
 \multicolumn{5}{|c|}{\textbf{2-dimensional, Variance = 0.5}} \\
 \hline
 \textbf{Algorithm}& \textbf{Accuracy} & \textbf{Recall} & \textbf{Precision} & \textbf{F Score} \\
 \hline
 Proposed k-means & 0.9705&0.9593&0.9625&0.9608\\
 \hline
 Traditional k-means& 0.9607&0.9445&0.9485&0.9464\\
 \hline
\end{tabular}
\caption{Results for 2D data with variance = 0.5}
\label{tab_2D_var_0.5}
\end{table}

The proposed \textit{k}-means algorithm that includes the ICD outperforms the traditional \textit{k}-means across the board. The Overall Accuracy, Recall, Precision and F-Score are all greater with the new proposed \textit{k}-means algorithm. One thing to note here is that the code is iterated 100 times and the scores in \textbf{table \ref{tab_2D_var_0.5}} represent the average over 100 iterations. 

\begin{figure}[H]
\centering
\includegraphics[width=1.25\textwidth, center]{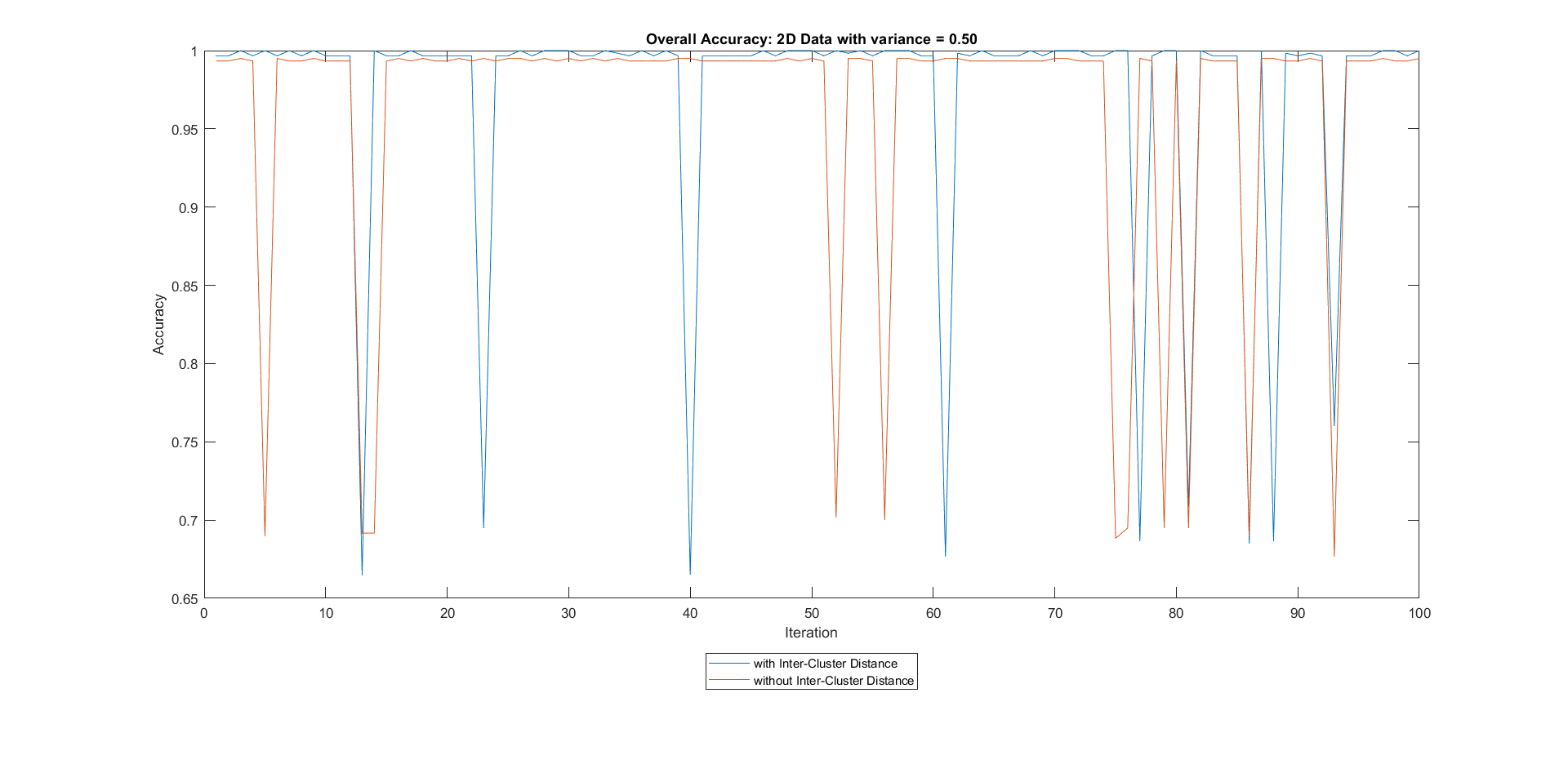}
\caption{Overall Accuracy of the proposed \textit{k}-means vs traditional \textit{k}-means over 100 iterations (variance = 0.5)}
\label{fig_OA_2D_var_0.5}
\end{figure}

We can see what the accuracy for each method looks like over 100 iterations in \textbf{figure \ref{fig_OA_2D_var_0.5}}. The faint blue line represents the new proposed algorithm while the orange line represents the traditional \textit{k}-means algorithm. It's clear that the results vary with every iteration with only a few consistent accuracy results. The new proposed algorithm reaches a perfect score numerous times throughout the 100 iterations while the traditional \textit{k}-means reaches the highest value of 0.9833 at its best. However, if noted closely, the new proposed algorithm has a lower accuracy during a few of those iterations. For example, during iteration 40, the new proposed algorithm drops below 70 percent accuracy. Running through the 100 iterations to make the results statistically viable also reveals a few weaknesses of the clustering algorithm itself as has been mentioned in multiple studies \cite{16}, \cite{17}, \cite{6}, \cite{5}. Since majority of the iterations range above 95 percent accuracy, the iterations that return less than that can be viewed as outliers. These iterations are the ones where the algorithm might have been susceptible to the initial centroid positions. One of the biggest weaknesses of \textit{k}-means is its sensitivity to the initial cluster centers. This could explain the iterations that return relatively lower accuracies. It is also interesting to note that through the 100 iterations, the cluster center initiation remains the same for both the proposed method and the traditional \textit{k}-means algorithm. This means that the permutation numbers that generate the random cluster centers are consistent for both methods. For example, if at iteration $25$ the initial cluster centers for one method were generated at \textit{x}, \textit{y}, \textit{z} coordinates of (1, 2, 3), then the initial cluster centers for the other method would also be at (1, 2, 3). Overall, the proposed algorithm evidently outperforms the traditional \textit{k}-means albeit by only 1 percent. This is a first indication of how including the ICD in the algorithm can generate better clustering results.

\subsection{2D Data with variance = 1}
\begin{figure}[H]
\centering
\includegraphics[scale=0.75]{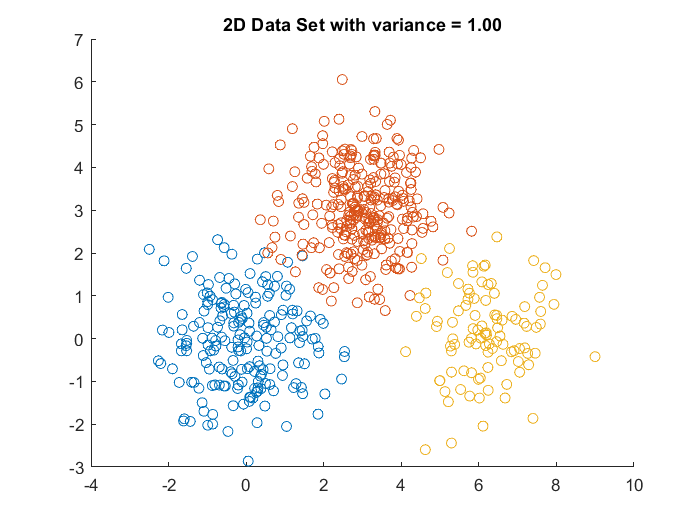}
\caption{2-dimensional data with variance = 1}
\label{fig_2D_var_1}
\end{figure}

\textbf{Figure \ref{fig_2D_var_1}} shows the 2-dimensional data set with a variance of 1. The cluster separation is less evident here as compared to a variance of 0.5 and yet the proposed algorithm outperforms the traditional \textit{k}-means by about 3 percent. The proposed \textit{k}-means returns an accuracy of 98.01 percent averaged over 100 iterations while the traditional \textit{k}-means returns 95.08 percent. Results for Recall, Precision and F Score are summarized in \textbf{table \ref{tab_2D_var_1}}. As outlined earlier during the literature research, most of the papers failed to provide a quantifiable method to evaluate the algorithms they proposed. One of the driving factors of this project was to be able to robustly quantify the proposed algorithm. 

\begin{table}[H]
\centering
\begin{tabular}[c]{ |p{2.5cm}||p{2.5cm}|p{2cm}|p{2cm}| p{2cm}| }
 \hline
 \multicolumn{5}{|c|}{\textbf{2-dimensional, Variance = 1}} \\
 \hline
 \textbf{Algorithm}& \textbf{Accuracy} & \textbf{Recall} & \textbf{Precision} & \textbf{F Score} \\
 \hline
 Proposed k-means & 0.9801&0.9816&0.9756&0.9785\\
 \hline
 Traditional k-means& 0.9508&0.9386&0.9400&0.9391\\
 \hline
\end{tabular}
\caption{Results for 2D data with variance = 1}
\label{tab_2D_var_1}
\end{table}

\begin{figure}[H]
\centering
\includegraphics[scale=0.75]{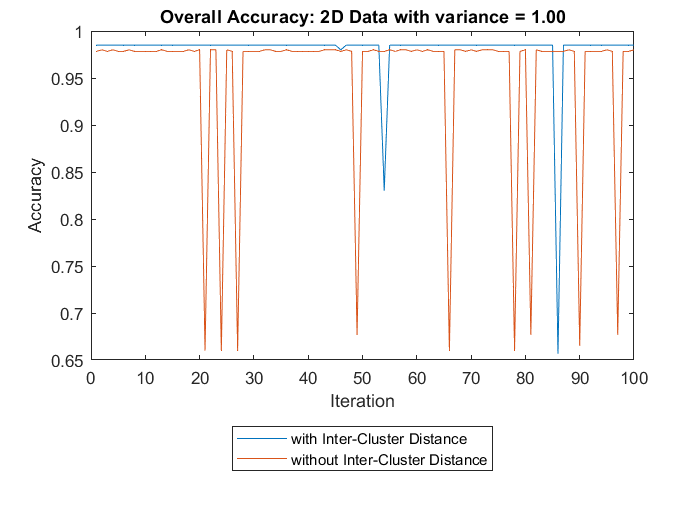}
\caption{Overall Accuracy of the proposed \textit{k}-means vs traditional \textit{k}-means over 100 iterations (variance = 1)}
\label{OA_fig_2D_var_1}
\end{figure}

We can see how often the traditional \textit{k}-means drops in accuracy to sub-70 percent as compared to the new proposed algorithm. This might be a result of the cluster center initialization but the code was set up in a way that had the same initialization positions for both methods. The overall accuracy is higher in the new proposed algorithm and stays more consistent than its traditional counterpart. Taking into account the average over 100 iterations, the new algorithm still outperforms the traditional one in terms of consistency and overall higher accuracy per iteration. 

\newpage

\subsection{3D Data with variance = 0.5}
\begin{figure}[H]
\centering
\includegraphics[width=1.25\textwidth, center]{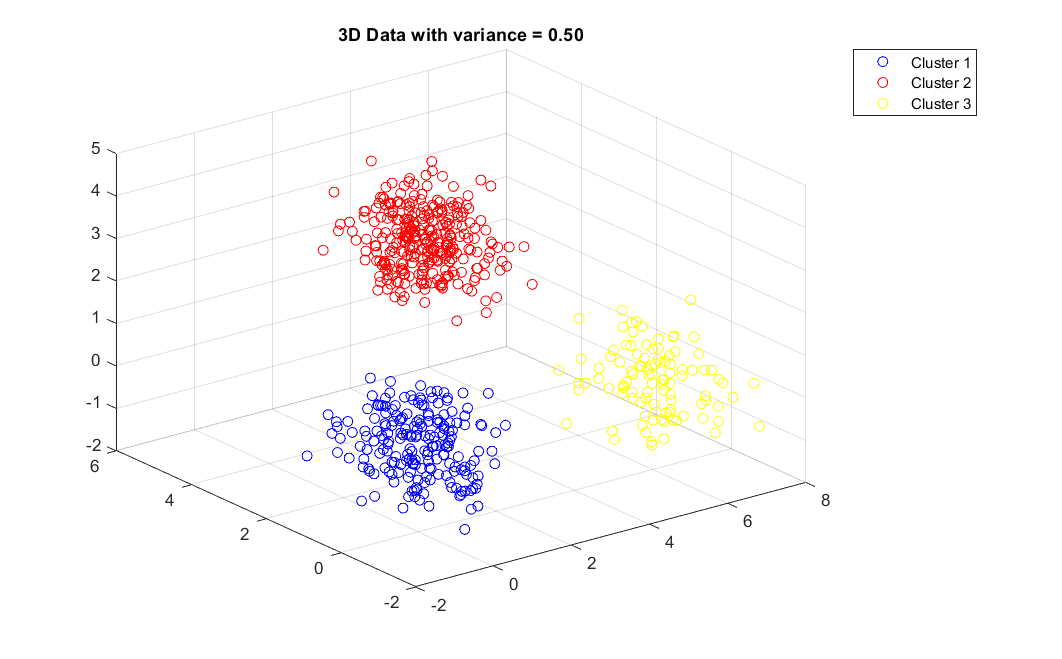}
\caption{3-dimensional data with variance = 0.5}
\label{fig_3D_var_0.5}
\end{figure}

The next synthetic data is a 3-dimensional data set with a variance of 0.5. You can see how the generated data is clustered in \textbf{figure \ref{fig_3D_var_0.5}}. 

\begin{figure}[H]
\centering
\includegraphics[width=1\textwidth, center]{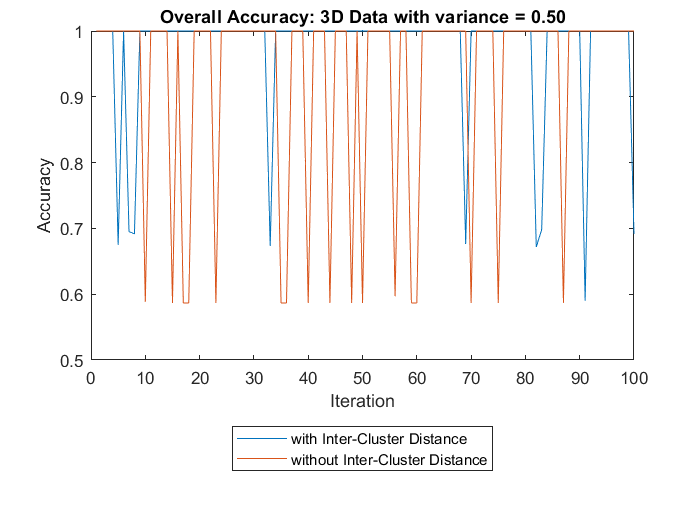}
\caption{Overall Accuracy of the proposed \textit{k}-means vs traditional \textit{k}-means over 100 iterations (variance = 0.5)}
\label{fig_OA_3D_var_0.5}
\end{figure}

The Overall Accuracy is just above 97 percent with the proposed algorithm and just under 93 percent for the traditional \textit{k}-means algorithm. The proposed algorithm outperforms the traditional method by about 4 percent. The Recall, Precision and F Score values are all tabulated in \textbf{table \ref{tab_3D_var_0.5}}. You can see the comparative performance of both methods in \textbf{figure \ref{fig_OA_3D_var_0.5}}. The instances where the initial cluster centers play a part in affecting the Overall Accuracy, the proposed method has fewer drops in accuracy as compared to the traditional \textit{k}-means. In addition, the drops are greater in magnitude for the traditional method. This contributes to the overall lower accuracy as we see in \textbf{table \ref{tab_3D_var_0.5}}.

\begin{table}[H]
\centering
\begin{tabular}[c]{ |p{2.5cm}||p{2.5cm}|p{2cm}|p{2cm}| p{2cm}| }
 \hline
 \multicolumn{5}{|c|}{\textbf{3-dimensional, Variance = 0.5}} \\
 \hline
 \textbf{Algorithm}& \textbf{Accuracy} & \textbf{Recall} & \textbf{Precision} & \textbf{F Score} \\
 \hline
 Proposed k-means & 0.9706&0.9566&0.9619&0.9591\\
 \hline
 Traditional k-means& 0.9298&0.9155&0.9239&0.9195\\
 \hline
\end{tabular}
\caption{Results for 3D data with variance = 0.5}
\label{tab_3D_var_0.5}
\end{table}

Analyzing the results in \textbf{figure \ref{fig_OA_3D_var_0.5}} further, we can infer that the traditional \textit{k}-means is more susceptible to the initial cluster centers as is evident through the iterations where the overall accuracy of the new proposed method is consistently perfect whereas the traditional \textit{k}-means drops to about 60 percent accuracy for the same iterations and the same initial cluster center positions. 

\newpage

\subsection{3D Data with variance = 1}
The results for the outputs are summarized in \textbf{table \ref{tab_3D_var_1}}. It is interesting to note that the Overall Accuracy is higher than the 3D data set with variance 0.5. This trend is also evident with the 2D data set. Results for a variance of 0.5 were lower than those for a variance of 1, even though the cluster separation becomes relatively ill defined as the variance increases.  

\begin{figure}[H]
\centering
\includegraphics[width=1.25\textwidth, center]{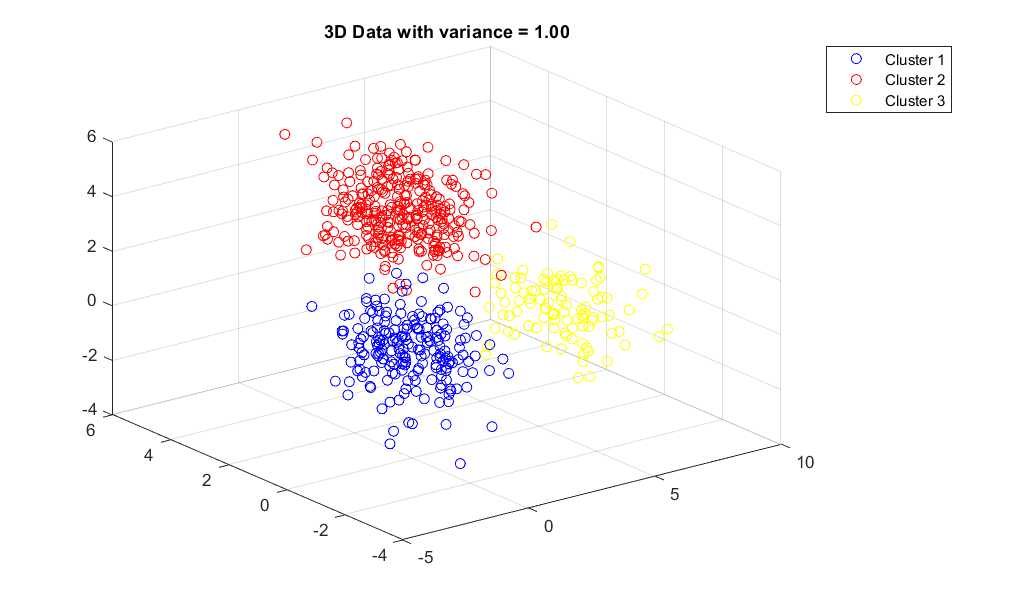}
\caption{3-dimensional data with variance = 1}
\label{fig_3D_var_1}
\end{figure}

\begin{table}[H]
\centering
\begin{tabular}[c]{ |p{2.5cm}||p{2.5cm}|p{2cm}|p{2cm}| p{2cm}| }
 \hline
 \multicolumn{5}{|c|}{\textbf{3-dimensional, Variance = 1}} \\
 \hline
 \textbf{Algorithm}& \textbf{Accuracy} & \textbf{Recall} & \textbf{Precision} & \textbf{F Score} \\
 \hline
 Proposed k-means & 0.9743&0.9755&0.9636&0.9695\\
 \hline
 Traditional k-means& 0.9514&0.9466&0.9422&0.9443\\
 \hline
\end{tabular}
\caption{Results for 3D data with variance = 1}
\label{tab_3D_var_1}
\end{table}

\begin{figure}[H]
\centering
\includegraphics[width=1\textwidth, center]{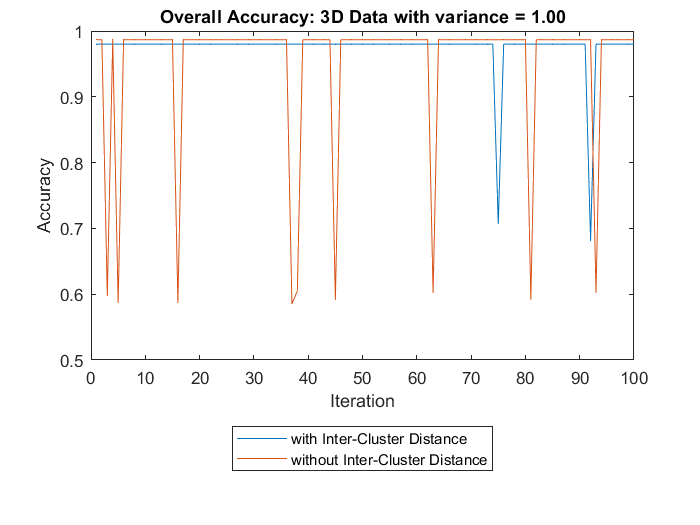}
\caption{Overall Accuracy of the proposed \textit{k}-means vs traditional \textit{k}-means over 100 iterations (variance = 1)}
\label{fig_OA_3D_var_1}
\end{figure}

The direct correlation between the variance and the overall accuracy of synthetic data set may also hint at the robustness of the algorithm and its ability to cluster relatively ambiguous data. You can see the evidence of this robustness in \textbf{figure \ref{fig_OA_3D_var_1}}. The overall accuracy of the proposed algorithm remains very consistent iteration over iteration while the traditional \textit{k}-means is sensitive to the initial cluster centers. The other thing to note is the magnitude of the drops in overall accuracy of the proposed algorithm versus the traditional methods. In iterations where the traditional \textit{k}-means drops to about 60 percent accuracy, the proposed algorithm stays around a near 100 percent accuracy with only a couple of drops. This could point to the two instances where the initial cluster centers might have contributed towards the poor clustering of the data. As is clear, even the proposed algorithm is not immune to the positions of the initial cluster centers. Regardless, the hypothesis of a better performance of the algorithm by including the ICD in the underlying \textit{k}-means algorithm has been solidified from the results of the four synthetic data sets we have presented so far. In the following sections, this hypothesis will be further tested using real-life benchmark data sets that were picked from the UCI Machine Learning Repository. 

\newpage
\section{UCI Machine Learning Repository}
\subsection{Iris Data set}
\begin{figure}[H]
\centering
\includegraphics[width=1.25\textwidth, center]{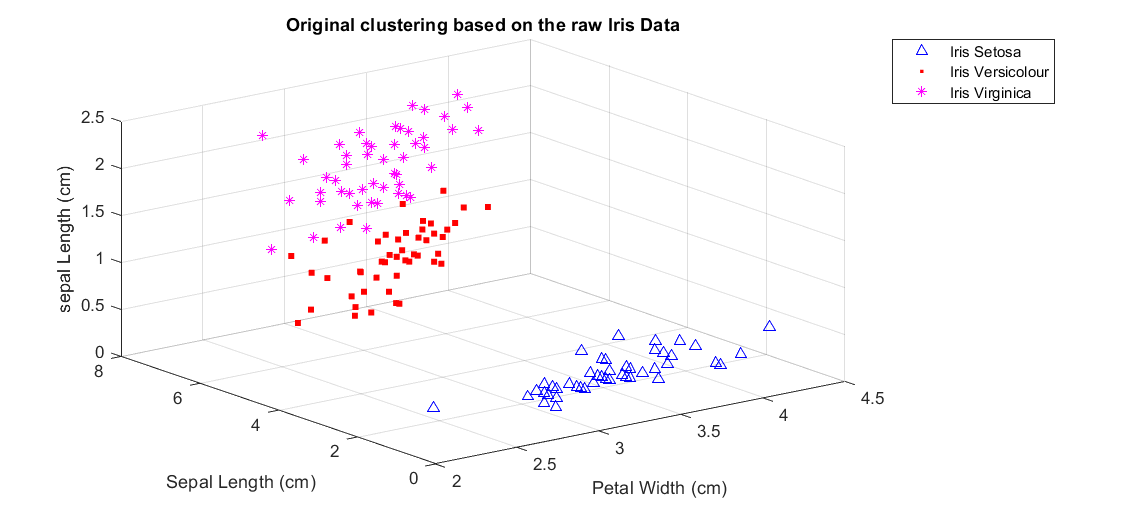}
\caption{The raw Iris data clusters}
\label{fig_Iris_clusters}
\end{figure}

The Iris data set is a benchmark data set for any kind of unsupervised machine learning algorithm analysis. It's the most widely researched data set. We can see how the data set is clustered in \textbf{figure \ref{fig_Iris_clusters}}. The three clusters represent the three types of Iris flowers: Setosa, Versicolour and Virginica. 

\begin{figure}[H]
\centering
\includegraphics[width=1.25\textwidth, center]{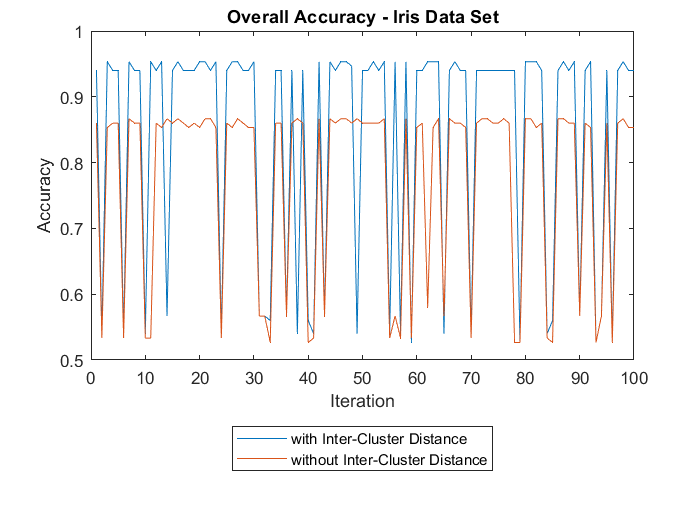}
\caption{Overall Accuracy for Iris}
\label{fig_OA_Iris}
\end{figure}

Running both the proposed algorithm and the traditional \textit{k}-means algorithm, we can see that the proposed algorithm outperforms the traditional method across the board. \textbf{Table \ref{tab_Iris}} shows the results from the run and \textbf{figure \ref{fig_OA_Iris}} shows the accuracy over 100 iterations.

\begin{table}[H]
\centering
\begin{tabular}[c]{ |p{2.5cm}||p{2.5cm}|p{2cm}|p{2cm}| p{2cm}| }
 \hline
 \multicolumn{5}{|c|}{\textbf{Iris Data set}} \\
 \hline
 \textbf{Algorithm}& \textbf{Accuracy} & \textbf{Recall} & \textbf{Precision} & \textbf{F Score} \\
 \hline
 Proposed k-means & 0.8420&0.8420&0.8347&0.8380\\
 \hline
 Traditional k-means& 0.7751&0.7751&0.7661&0.7703\\
 \hline
\end{tabular}
\caption{Results for Iris Data}
\label{tab_Iris}
\end{table}

What is interesting to note is that even though the drops in accuracy at certain intervals align almost perfectly between the two methods, the maximum accuracy achieved through the proposed algorithm is higher than the maximum accuracy with the traditional method. The proposed algorithm reaches a maximum accuracy of about 95.33 percent while the traditional method cannot break the 85 percent barrier. The drawback of the initial cluster centers is evident in both methods but it weighs more on the traditional \textit{k}-means than it does on the proposed algorithm pointing to the advantage of including the ICD. While the accuracy for the new proposed method is not as high as some of the literature studies have achieved, it is evidently higher than the traditional \textit{k}-means algorithm, which is what was hypothesized for this project. One study \cite{18} achieves a 92.67 percent accuracy using a genetic algorithm with the traditional \textit{k}-means. Another study \cite{20} achieves a 95 percent and a 99 percent accuracy with the two methods they propose in their study that happens to address the issue of the sensitivity to the initial cluster centers.

\begin{figure}[H]
\centering
\includegraphics[width=1.25\textwidth, center]{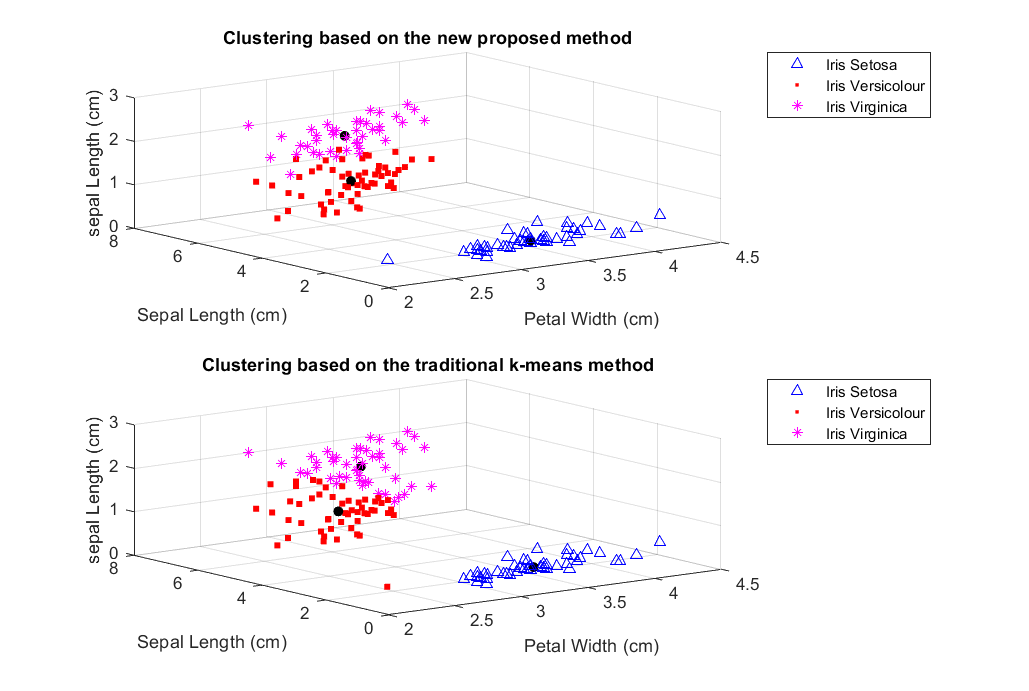}
\caption{Clusters produced by the proposed method vs traditional \textit{k}-means}
\label{fig_Iris_prop_trad_custers}
\end{figure}

\textbf{Figure \ref{fig_Iris_prop_trad_custers}} visually compares the clustering achieved by the two methods. The upper subplot is the clustering for the proposed method and the lower subplot is for the traditional method. Both plots also have the cluster centers to better be able to understand how the algorithm works. There is a outlier data point near the coordinate (2,2,0) that should be clustered as Iris Setosa but the traditional method fails to do so. The proposed method correctly identifies this data point. Upon taking a closer look, the answer may be in the cluster centers themselves. As has been mentioned multiple times through this paper, the algorithm is very susceptible to the initial cluster centers. You can see that the cluster center for the Versicolor is somewhat in the background for the proposed method as opposed to the relatively foreground position for the traditional method. Since we are choosing new cluster centers by averaging the within cluster distance at every iteration, including the ICD may provide a more robust clustering as supposed to the traditional \textit{k}-means method. The cluster center being positioned in the background forces the algorithm to cluster the data point correctly as opposed to missing it in the traditional method. 

%%%%%%% WINE DATA %%%%%%%
\subsection{Wine Data set}
\begin{figure}[H]
\centering
\includegraphics[width=1.3\textwidth, center]{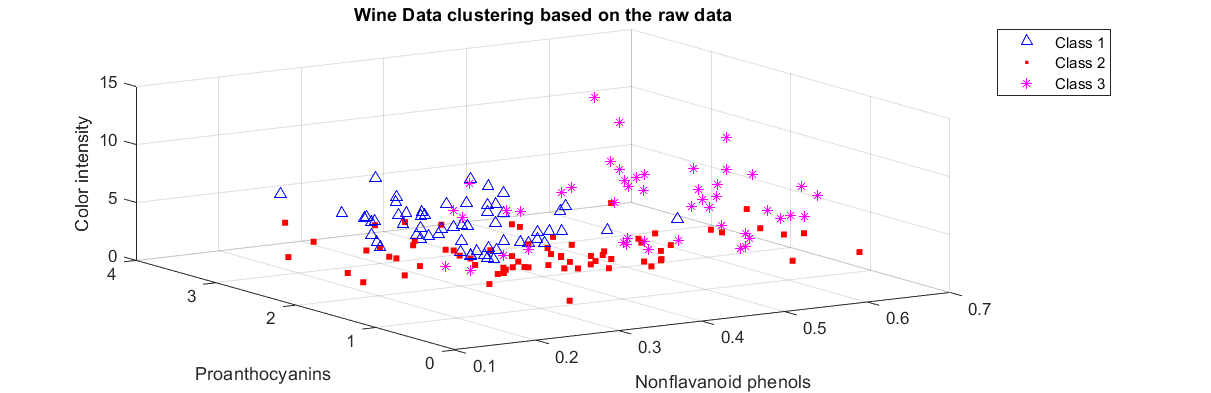}
\caption{The raw Wine data clusters}
\label{fig_Wine_clusters}
\end{figure}

Results of the wine data set are summarized in \textbf{table \ref{tab_Wine}}. It is important to note that the scatter plot shown in \textbf{figure \ref{fig_Wine_clusters}} only shows 3 of the 13 dimensions or attributes of the entire data set. The proposed algorithm resulted in a higher overall accuracy along with recall, precision and F scores for the data as compared to the traditional method. Although the accuracy is not very high, resulting in only a 75.16 percent accuracy, it outperforms the traditional method by almost 9 percent. As a reminder, the goal of this project was not to aim for a very high accuracy but rather to create more accurate clusters of the data. Both methods were ran 100 times and the scores in \textbf{table \ref{tab_Wine}} reflect the average over 100 iterations. You can see how consistent the proposed algorithm is with very minimal susceptibility to initial cluster centers while the traditional \textit{k}-means being very volatile. This volatility can be attributed to the sensitivity of the traditional method to the initial cluster centers.
\begin{table}[H]
\centering
\begin{tabular}[c]{ |p{2.5cm}||p{2.5cm}|p{2cm}|p{2cm}| p{2cm}| }
 \hline
 \multicolumn{5}{|c|}{\textbf{Wine Data set}} \\
 \hline
 \textbf{Algorithm}& \textbf{Accuracy} & \textbf{Recall} & \textbf{Precision} & \textbf{F Score} \\
 \hline
 Proposed k-means & 0.7516&0.7318&0.7609&0.7460\\
 \hline
 Traditional k-means& 0.6637&0.6698&0.6863&0.6778\\
 \hline
\end{tabular}
\caption{Results for Wine Data}
\label{tab_Wine}
\end{table}

\begin{figure}[H]
\centering
\includegraphics[width=1\textwidth, center]{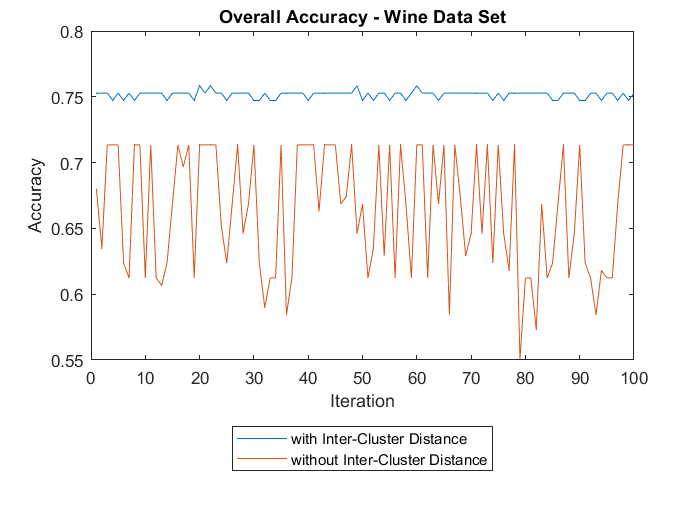}
\caption{Overall Accuracy for Wine}
\label{fig_OA_Wine}
\end{figure}
\begin{figure}[H]
\centering
\includegraphics[width=1.25\textwidth, center]{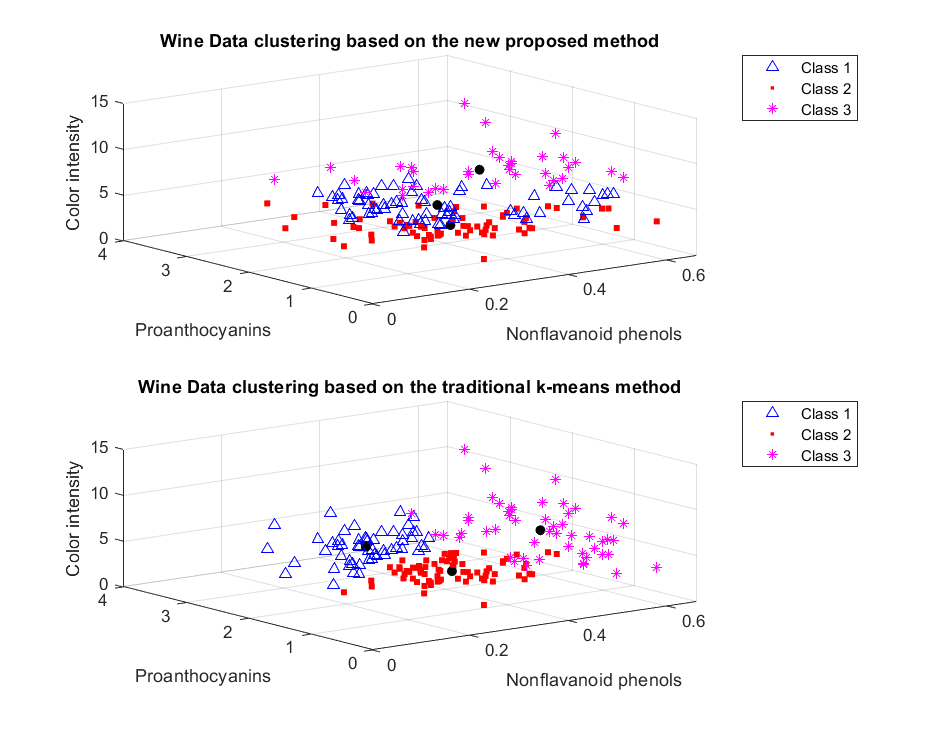}
\caption{Clusters produced by the proposed method vs traditional \textit{k}-means}
\label{fig_Wine_prop_trad_custers}
\end{figure}

\textbf{Figure \ref{fig_Wine_clusters}} shows us the ambiguity in the cluster separation. Despite this ambiguity, the proposed algorithm was able to not only perform better overall, but it also had a lower sensitivity to the cluster centers and hence a less volatile accuracy output. To better understand the difference in both algorithms, \textbf{figure \ref{fig_Wine_prop_trad_custers}} plots the clustered data forced by the algorithms along with the cluster centers. On first glance, it may seem like the traditional \textit{k}-means has relatively clearer separation in its clustered data, but the accuracy of those clusters is very low compared the clusters formed from the proposed algorithms. It is also interesting to note how robustly the proposed algorithm correctly classifies the outliers in the data. Outliers here do not mean outliers from a statistical perspective but rather they signify the data points that are further away in distance than the majority of the data points in the clusters. In this figure, the points at coordinates of about (0.6, 0.5, 2.5) and (0.3, 3.5, 2.5) have been correctly identified as Class 2 by the proposed algorithm but the same have been misclassified by the traditional method. We saw the same trend with the Iris data set. It is becoming clear that the proposed algorithm is robust enough to correctly classify outliers. Albeit, we have only seen this trend with two data sets so the statistical significance of this observation is pretty low. Nonetheless, we have seen consistent outperformance in accuracy of the proposed algorithm over the traditional method.

%%%%%% BREAST CANCER DATA %%%%%%%%%
\subsection{Breast Cancer Data set}

\begin{figure}[H]
\centering
\includegraphics[width=1\textwidth, center]{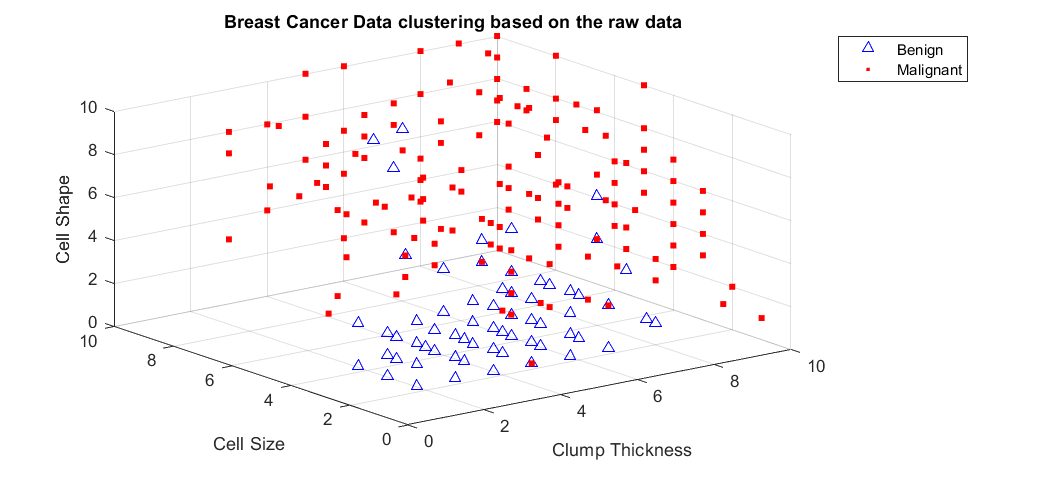}
\caption{The raw breast cancer data set clusters}
\label{fig_breast_clusters}
\end{figure}
The breast cancer data set comprises of the digital information from images of breast mass, describing the characteristics of cell nuclei present in the image. The data set itself is a 9 dimensional data set with the last feature being the  diagnosis (classification) of that data point (either Benign or Malignant). All the previous data sets we've used have only been either a 2-dimensional or a 3-dimensional data set because the proposed algorithm itself was synthesized for lower dimensional data with a gaussian distribution. We hypothesized that including the ICD in the k-means algorithm results in higher overall accuracy and more robust clustering in the data. That hypothesis has been proven with the results of all the data sets so far. For our final experiment, it was interesting to stress-test the algorithm with a higher dimensional data set and observe its performance with regards to the same metrics used to quantify the previous results - Overall Accuracy, Recall, Precision and F Score. As shown in \textbf{table \ref{tab_breast}}, the difference between the two methods is minuscule. Overall Accuracy only drops about 0.3 percent and the same difference in numbers trickles down to the other metrics. Maybe this small change is a result of the data being more than 3 dimensions as has been the case in all the previous data sets. Despite the small jump, the proposed algorithm manages to outperform the traditional method.

\begin{table}[H]
\centering
\begin{tabular}[c]{ |p{2.5cm}||p{2.5cm}|p{2cm}|p{2cm}| p{2cm}| }
 \hline
 \multicolumn{5}{|c|}{\textbf{Breast Caner data set}} \\
 \hline
 \textbf{Algorithm}& \textbf{Accuracy} & \textbf{Recall} & \textbf{Precision} & \textbf{F Score} \\
 \hline
 Proposed k-means & 0.9612&0.9532&0.9611&0.9571\\
 \hline
 Traditional k-means& 0.9582&0.9499&0.9578&0.9538\\
 \hline
\end{tabular}
\caption{Results for Breast Data set} 
\label{tab_breast}
\end{table}

\begin{figure}[H]
\centering
\includegraphics[width=1.2\textwidth, center]{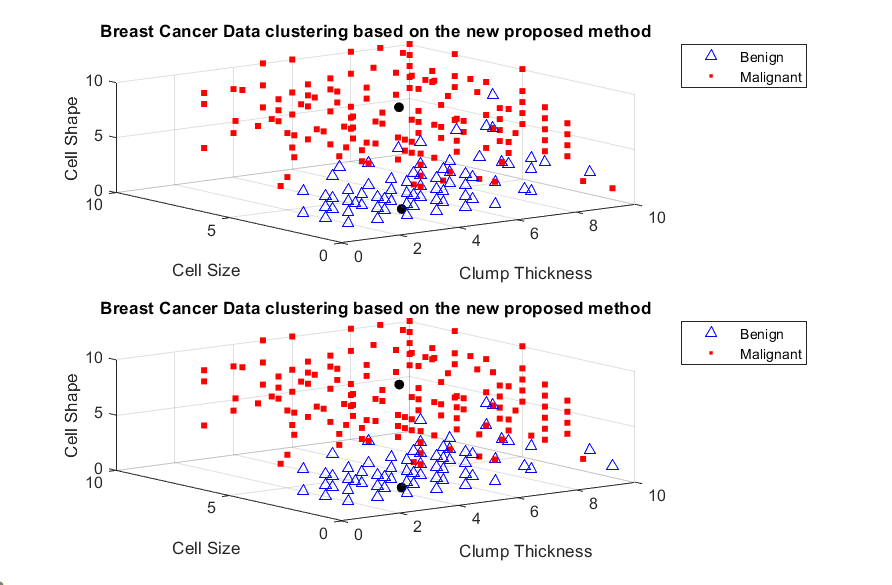}
\caption{Clusters produced by the proposed method vs traditional \textit{k}-means}
\label{fig_breast_prop_trad_clusters}
\end{figure}

The clusters produced by the two methods are shown in \textbf{figure \ref{fig_breast_prop_trad_clusters}} and it is obvious that both methods are pretty close in their performances. The cluster centers from the last iteration that are shown, are almost identical for both methods and the clustering itself has a significant overlap between the two except for a few of the misclassified data points as is expected. The algorithm handled the stress test fairly well despite being catered towards a lower dimensional data set. In addition, the proposed method is less volatile in its performance over the 100 iterations as compared to the traditional method. To validate our evaluation metrics, let's say that in an extreme case the clustering caused all data points to move to one cluster resulting in a class imbalance. At its worst, the overall accuracy would still only fall to about 50 percent, which checks out from a hypothetical perspective in terms of the mathematics behind the evaluation. 

\begin{figure}[H]
\centering
\includegraphics[width=1.2\textwidth, center]{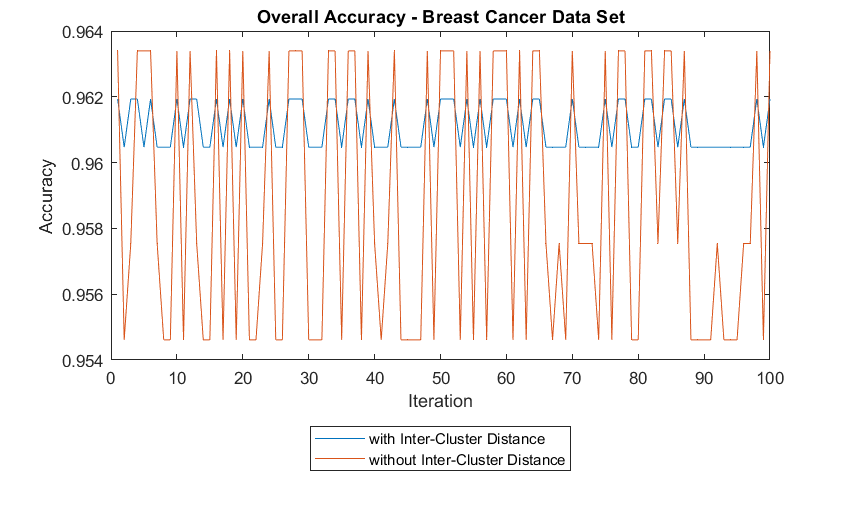}
\caption{Overall Accuracy for Breast Cancer}
\label{fig_OA_breast}
\end{figure}

We can see in \textbf{figure \ref{fig_OA_breast}} how the volatility compares between the two methods, iteration over iteration. We've seen this happen with the Iris and Wine data sets as well but with the breast cancer data set, it is more robust and apparent. While the proposed method consistently hovers in the 96.2 to 96 percent range, the traditional method jumps between 96.4 to 95.4 percent. Although it is worth noting that the traditional method does score higher than the proposed method. The sensitivity to the initial cluster centers is definitely a factor worth considering when choosing an appropriate algorithm to deploy for a data set. Observations over the data sets used in this project have clearly shown that, albeit in small numbers. It can be a launching pad for research behind the nuanced numerical methods involved in k-methods clustering and clustering analysis in general to see if there are more efficient ways to avert the sensitivity to initial cluster centers. A large number of papers that were read for the literature review on this project suggested multiple initialization techniques and also distance measures that would prove beneficial for a novel approach with high accuracy measures \cite{5}, \cite{6}, \cite{7}, \cite{33}.

\chapter{Conclusion}
In this project, we have demonstrated the superiority of using the inter-cluster distance in the \textit{k}-means clustering. By including the ICD along with the WCD, the new proposed algorithm was able to outperform the traditional \textit{k}-means across a variety of data sets with different dimensions and clusters. It is widely known that \textit{k}-means clustering is a popular unsupervised clustering method that is heavily reliant on the prior knowledge of the number of clusters in a data set. The evaluation method for determining the optimum number of clusters was the Calinski-Harabazs index and the elbow method as a follow up to confirm the findings of the Calinski-Harabazs method. Traditional \textit{k}-means is also extremely sensitive to the initial clusters and its accuracy is highly dependent on how far the initial cluster centers are from the true cluster centers. There were obvious implications that were evident in the accuracy scores in some of the data sets. We were able to address and analyze the effects of the same in this project.
\par
We were also able to stress-test the algorithm with a high dimensional data to confirm its validity. Initially the synthesis of this proposed algorithm was based on a 2 or 3-dimensional data with a gaussian distribution but by testing the algorithm with a high dimensional data set, we were able to see that the proposed algorithm is robust enough to cluster high dimensional data with a higher accuracy than its traditional counterpart. We also saw a trend with the synthetic data sets where the data sets with a higher variance had a higher overall accuracy. The validity of this claim is yet to be explored in its full capacity but this project definitely sets up the precedent for the same. Overall, the proposed algorithm outperformed the traditional \textit{k}-means across all the data sets that were used in this research. There were instances where the overall accuracy was higher by a small amount but nonetheless, there were ample of evidences to prove our initial hypothesis correct. 
\par A number of studies also evaluated the optimization of the final clusters based on the inter-cluster distance and all of them also came to the same conclusion - the inclusion of another distance metric definitely boosts the accuracy of the traditional \textit{k}-means algorithm. Although our approach with the algorithm in this project was a little different than the rest, the results were in conjunction with the rest of the published work.

\chapter{Future Work}
\textit{k}-means is probably the most widely researched topic in clustering analysis. Unsupervised learning is a deep rabbit hole with nuanced approaches that is addressed in a lot of published papers. Deep learning has been the hot topic for the past several years that researchers seem excited about. In this pursuit of higher accuracy clustering results, \textit{k}-means certainly lends itself to set up more interesting research in the field of deep learning. Machine learning paths like computer vision, natural language processing, predictive analysis and operations research all have some form of clustering analysis associated with them.
There are few ways this project sets up work for future research:
\begin{itemize}
    \item The effect of variance on the accuracy of the proposed algorithm.
    \item The validity of the proposed algorithm on high dimensional data.
    \item Cluster center initialization methods and their effect on the overall accuracy.
    \item Using an optimization technique to address the sensitivity to the initial cluster centers.
\end{itemize}

There are many other avenues one can pursue and build upon, based on this project. The hope is that the work done in this project provides motivation and material for more sophisticated approaches towards \textit{k}-means research.

\listoffigures
\listoftables

\printbibliography

\end{document}